\documentclass[runningheads]{llncs}
\usepackage{graphicx}
\usepackage{xcolor}
\usepackage{multirow}
\usepackage{float}
\usepackage{booktabs}
\usepackage{color}
\usepackage{colortbl}
\begin{document}
\title{Cyrus2D base: Source Code Base for RoboCup 2D Soccer Simulation League}
%
\author{
Nader Zare\inst{1}\and 
Omid Amini\inst{4}\and
Aref Sayareh\inst{5}\and
Mahtab Sarvmaili\inst{1}\and
Arad Firouzkouhi\inst{6}\and
Saba Ramezani Rad\inst{6}\and
Stan Matwin\inst{1}\inst{2} \and
Amilcar Soares\inst{3}}
\authorrunning{N. Zare et al.}
%
\institute{
Institute for Big Data Analytics, Dalhousie University, Halifax, Canada\\
\and
Institute for Computer Science, Polish Academy of Sciences, Warsaw, Poland\\
\and
Memorial University of Newfoundland, St. John's, Canada\\
\and
Qom University of Technology, Iran\\
\and
Shiraz University, Iran\\
\and
Amirkabir University of Technology, Iran\\
\email{\{nader.zare, mahtab.sarvmaili\}@dal.ca},\\ 
\email{\{arefsayareh, omidamini360\}@gmail.com},\\
\email{\{arad.firouzkouhi, saba\_ramezani\}@aut.ac.ir}\\ 
\email{stan@cs.dal.ca}, \email{amilcarsj@mun.ca}
}
\maketitle              
\begin{abstract}
Soccer Simulation 2D League is one of the major leagues of RoboCup competitions. In a Soccer Simulation 2D (SS2D) game, two teams of 11 players and one coach compete against each other. Several base codes have been released for the RoboCup soccer simulation 2D (RCSS2D) community that have promoted the application of multi-agent and AI algorithms in this field. In this paper, we introduce "Cyrus2D Base", which is derived from the base code of the RCSS2D 2021 champion. We merged Gliders2D base V2.6 with the newest version of the Helios base. We applied several features of Cyrus2021 to improve the performance and capabilities of this base alongside a Data Extractor to facilitate the implementation of machine learning in the field. We have tested this base code in different teams and scenarios, and the obtained results demonstrate significant improvements in the defensive and offensive strategy of the team.


\footnote{NOTICE: This is an accepted article to be published by Lecture Notes
in Artificial Intelligence (LNCS/LNAI) series by Springer-Verlag in the
proceedings of the RoboCup International Symposium.}
\keywords{2D Soccer Simulation \and RoboCup \and Base Code.}
\end{abstract}
\section{Introduction}
Soccer is one of the most popular team-based sports in the world. This is a multi-player, real-time, strategic, and partially observable game in which players of each team should cooperate to score more goals. 
In addition to the cooperative strategy, the players should manage different tactical and technical strategies against their opponent. 
Designing and implementing this game in a good, realistic graphical simulation environment and encouraging researchers to develop fully autonomous players with human-like skills creates complex challenges for A.I. research.
Hence, soccer is considered an exciting environment for developing A.I. and robotic algorithms to solve real-world challenges. 
The importance of soccer as a game and as a challenging domain for testing the A.I. and machine learning algorithms led to an overreaching vision of a robotic team competing against the best human team by 2050\cite{rc50}. 

The world Cup Robot Soccer Initiative was founded to create a realistic environment similar to real soccer that encourages researchers to employ Robotic and A.I. for solving wide ranges of problems\cite{noda1996soccer}.  The first RoboCup was held during the IJCAI-97 \cite{robo1997}, and it offered three competition tracks: real robot league, software robots, and expert robot competition. Among them, the Soccer Simulation 2D league (SS2D) \cite{kitano1997robocup} provides a wide range of research challenges such as autonomous decision-making, communication and coordination, tactical planning, collective behaviour and teamwork, opponent modelling and behavior predicting \cite{ref1,ref2,ref3,ref4,ref5,ref6,ref7,cyrussamposiom}.

\begin{figure}[ht]
    \centering
    \includegraphics[scale=0.250]{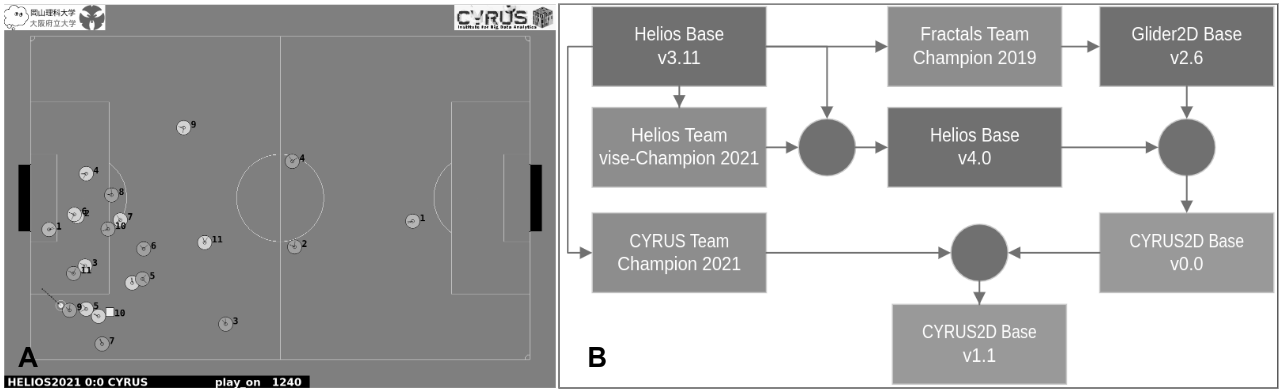} \\
    \caption{A: Soccer Simulation 2D League. B: The evolution of Helios2D, Glider2D and Cyrus2D base codes}
    \label{fig:2d}
\end{figure}

In this league, the RoboCup Soccer Simulation Server (RCSSServer) executes and manages a 2D soccer game between two teams of eleven autonomous software programs(agents).
It holds the complete knowledge of the game, such as the exact position of every element in the game and their movements. The game further relies on the communication between the server and each agent. Each player receives relative and noisy information about the environment, and based on its logic and algorithms, the agent produces basic commands (like dashing, turning, or kicking) to influence the environment. A visual example of the game is shown in Figure \ref{fig:2d}. 
Another key component of this league is the base code\footnote{For simplicity , throughout this paper we will use the "base" term instead of base code.} of agents that is responsible for communicating with the server, handling the noisy partial observability of the game, modeling the server world, and making multi-agent decisions throughout the game. Due to the complexity of these tasks, designing an operational base code can astonishingly accelerates the RSS2D teams' progress.    

Over the past years, many teams have contributed to the RCSS2D community by releasing their bases, which are mentioned below.
One of the first bases was from Carnegie Mellon University, a.k.a "CMUnited" in 2001 \cite{CMR0,CMR1}, then a windows-based team was released by "TsinghuAeolus" \cite{TAR} in 2002. The release of "UvA Trilearn" base \cite{UTR} in 2003, helped many teams worldwide.
"Brainstormers" \cite{BSR}, "WrightEagle" \cite{WRR} and "Marlik" \cite{MKR} released their team codes in 2005, 2011 and 2012 respectively. "HELIOS-Base or Agent2D" has been released by the "HELIOS" team from AIST Information Technology Research Institute \cite{agent2d,ag3}; this is the most important, most relevant, and most frequently used publicly available source code release in soccer simulation 2d.
It has been considered as the base code for many prosperous teams such as Cyrus2d \cite{cyrus13,cyrus21,cyruschampion} and Glider2d\cite{gld2016,gld2019}.  Later, "Cyrus2D" and "Gliders2D" released their 2014 \cite{cyrus14} and 2019 \cite{glbase} bases respectively that are based on agent2d.




In this paper we are planning to describe and release a more advanced base that is called Cyrus2D in three consecutive versions. We have followed the incremental strategy of evolving base code proposed by \cite{glbase,glbase2} to exemplify the impact of different approaches and to trace their functionalities.  
In the first version(v0.0) we combined the newest release of Helios and Gliders bases with some modifications on their parameters. In the second version(v1.0), we have developed this base code to include three A.I.-based components of Cyrus2D  that were successfully implemented in this team. Finally, in the third version(v1.1) we took advantage of Pass Prediction Deep Neural Network module for unmarking decision-making. The performance of these versions went on the rigorous evaluations against the Agent2D, Glider2D bases and the obtained outcomes (number of scored and received goals, and the winning rate) proved the prevalence of our base code. The rest of the paper is organized as follows:
Section 2 we will define the foundation of our base (version zero), in the next section we will explain the deployment of three ideas(Blocking Strategy, Offensive Risk Evaluation, Simple Unmarking Strategy) on the Cyrus2D v0.0 which results in Cyrus2D v1.0. In section 4, we will present the idea of using Pass Prediction in Unmarking Strategy(Cyrus2D v1.1). In the next section, we will compare Cyrus2D base with other Soccer Simulation 2D bases against best three teams in RoboCup 2021.
Finally, we talk about our future works.

\section{Cyrus2D Base Version 0.0}
One of the most popular SS2D bases is the Helios Base (agent2d) V3.11 which was released in 2010\cite{agent2d,ag3}. This base includes several components such as \textit{librcsc-4.0.0, soccerwindow2-5.0.0 and fedit2-2.0.0}. Gliders and Fractals, who use the Agent2D base, won the championship of RoboCup 2016 and 2019, respectively\cite{gld2016,gld20162,gld2019}. They also released a simplified version of their teams called Gliders2d base\cite{glbase,glbase2}. It is an advanced version of Helios base v3.11 with improved formation, passing behavior, and stamina management. It employs a modified version of the Marlik team\cite{MKR} blocking algorithm, and few unique strategies specifically designed for each team.
On the other hand, Helios has started improving its base and components such as \textit{librcsc} based on the new versions of C++ from 2019\cite{agnew,libnew}. 
In this paper we start by introducing the first version of Cyrus2D base (V0.0). It is established by rewriting the newest version of the Agent2D by merging the latest Gliders2D base (see Figure\ref{fig:2d}[B]). This base code is fully compatible with the latest version of \textit{librcsc}, but the blocking algorithm and tuning parameters of the Gliders2D base are removed. The Cyrus2D base is released in the Cyrus team repository and will be updated to be compatible with the \textit{rcssserver} and \textit{librcsc}\footnote{https://github.com/Cyrus2D/Cyrus2DBase}. In order to enhance the functionality of this base, we have implanted three simplified functionalities of Cyrus on this base and we introduce them as the consecutive versions of Cyrus2D base. In the following sections, we will describe these ideas.

\section{Cyrus2D Base Version 1.0}
\subsection{Blocking Strategy \protect\footnote{This algorithm is implemented in src/bhv\_basic\_block.cpp} }
As the environment of SS2D is highly dynamic and unpredictable, an innovative defensive strategy can increase the winning chance of the team. To establish the defensive strategies, we need to understand defensive actions and how players can cooperatively perform to minimize the risk of receiving the goal.      
Blocking and marking are two main defensive actions that prevent the opposing team from controlling the ball and playing with it. Blocking stops the progress of the opponent's ball holder on the field, and marking prevents the passing of the ball to the opposing team players.
Therefore, when one of our agents tries to block the ball holder, the other players should choose to mark the opponent players. 
In the Cyrus2D base, we implemented multi-agent blocking decision-making. The blocking function or "Blocking Simulator" is called when the opponent owns the ball. It simulates the dribbling behavior of the opponent ball holder called the "dribbling curve" and then finds a position that one of our players can arrive in, before arrival of the opponent's ball holder and (our) players. To simulate the dribbling curve, it predicts the first position of the ball that the opponent's player can kick the ball.
In the next step, it predicts the following ball positions of dribbling behavior. The dribbling speed is considered 0.7 m/s. To find the dribbling direction, we evaluate ten positions around the ball position using the reversed formulation of "Field Evaluator" in Helios base.
To improve the performance of the Blocking algorithm, we implemented some conditions to prevent players from using extra stamina or going far from their home position.

\subsection{Offensive Risk Evaluation \protect\footnote{This algorithm is implemented in src/chain\_action/action\_chain\_graph.cpp }}
To score more goals, the team's ball holder must move the ball towards the opponent's goal area, and a final striker must shoot the ball towards the goal.
Dribbling and passing are examples of possible actions that can lead the ball towards the goal. Henceforth, the ball holder must choose the best action between the possible passes and dribbles. For this purpose, we need to scrutinize our base code and improve the implementation of the offensive strategy.

The Agent2D base has a decision-making algorithm called \emph{Chain-Action}, which uses a modified version of Breadth-First Search to decide an action for the ball owner in an action graph tree. 
The Chain-Action has \emph{action generator} modules such as Pass-Generator, Short-Generator, and Dribble-Generator. An \emph{action generator} module receives a state of the game and then generates all possible actions in that state. The Chain-Action also includes a simple \emph{predictor module} that receives a state and an action; then, it generates a new state. It simulates the possible outcome of the game after applying the received action\cite{cyruschampion}.

\begin{figure}[ht]
    \centering
    \includegraphics[width=.5\textwidth]{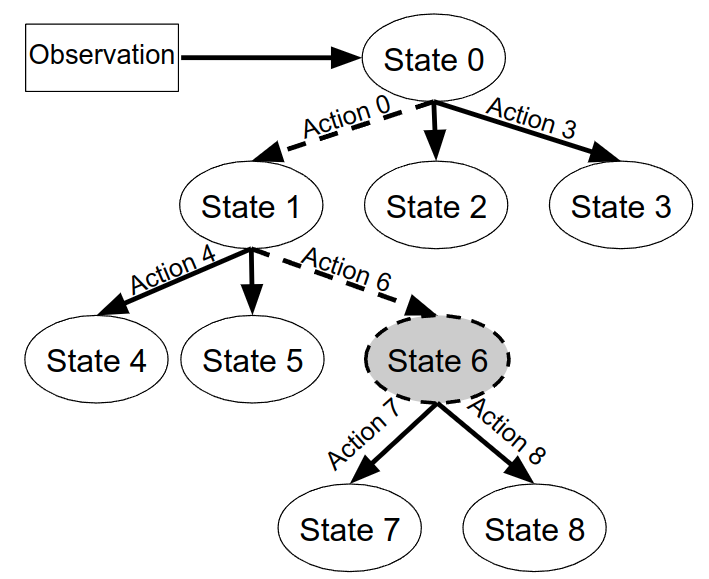} 
    \caption{Example of Chain-Action. A multi-branch tree search is performed. Each edge presents an action and each node corresponds to a state instance.}
    \label{fig:chain}
\end{figure}

After predicting a new state, Chain-Action evaluates the state based on the ball position using a module called \emph{Field-Evaluator}. This module receives a state and uses the $X$ coordinate of the ball and its distance to the opponent's goal to measure its value. 
To expand the tree to the next level, the chain-action chooses a node with the maximum value. An example of this procedure is shown in Figure \ref{fig:chain}.




We improved the Field-Evaluator module by including a term that is subtracted to its calculation algorithms which is called \emph{Offensive Risk Evaluation (ORE)}. This value is calculated based on the minimum number of cycles that the opponent players need to reach the ball in the input state of Field-Evaluator.

The Field-Evaluator first calculates the minimum number of cycles $c$ that the opponent player needs to reach the ball. Then it uses an array with seven elements where the n-Th element would be the ORE term if the opponent reaches the ball in n-Th-cycles. To populate this array, we took advantage of the genetic algorithm\footnote{We reduced the array size to seven because our GA algorithm with several settings found that the eighth and following cells of the best arrays will be 0.}.

For our task, the genetic representation is a list of seven values. A solution must be an array of seven values between 0 to 50, in descending order, as opponents closer to the ball are more dangerous.
The fitness function is the average goal difference of Cyrus2D base in 100 games against random opponents from 10 teams of RoboCup competitions in 2021.
To initialize the first population, we randomly generated 100 solutions.
After evaluating with the fitness function, we generate 80 new children, from 160 parents which are selected randomly with probability based on their fitness score. After cross-over, we update the new children to possible solutions by making sure each value is less than or equal to the value before it.
In the next step, some of their genes can be mutated with a low random probability. The mutation is done in a manner that the mutated solution is still considered possible.

If the generated solution is not in descending order, to preserve the validity of the solution, we replace the first occurrence of illegal value with a value lower than the previous element for example if the generated solution looks like \textit{solution} = [10 18 5 4 3 2 1], the validity procedure transforms it to \textit{fixed solution} = [10 9 5 4 3 2 1].
Afterwards, we create the new population of $100$ by selecting $20$ of the best chromosomes of the previous generation and adding the $80$ new children.

We repeat this process until the population converges or until $100$ iterations are evaluated. 

\subsection{Unmarking Strategy \protect\footnote{This algorithm is implemented in src/bhv\_unmark.cpp}}
Unmarking is the player's ability to move, avoid being marked, and relocate himself in a space where he could receive a pass from the ball possessor.

In the unmarking algorithm, a player who wants to unmark is called the "unmarker", and the player who will pass the ball to the unmarker is called the "passer". The passer player can be a player who owns the ball or does not have ball possession at the moment, but it is possible to be a ball possessor in the future. An unmarking Strategy identifies the passer, and after identifying the passer, the unmarker should find a position to receive a pass from the passer in future cycles.
An effective unmarking should consider the actions of other agents and the cooperation between them. 

Cyrus2D base version 1.0 includes a simple unmarking strategy. In this algorithm, all players do unmarking for the ball possessor to receive a pass from him. After identifying the passer, the unmarker simulates ten targets in ten directions around him according to its previous movement. 
After generating 100 targets, it ignores targets close to teammate or opponent players and targets far from its home position. The home position is the target position of a player that is calculated based on team's formation.
The next step simulates eight lead passes from the passer to itself in every target to find which target it can receive a pass. A pass has a score calculated using the "Field Evaluation" formula in the Helios base. The score of each target is calculated based on the scores of possible received passes in the position and the minimum distance of opponent players to the position. Eventually, the target with the maximum score will be selected as the unmarking target.

\section{Cyrus2D Base Version 1.1 \protect\footnote{This algorithm is implemented in src/bhv\_unmark.cpp, src/data\_extractor/DEState.cpp and src/data\_extractor/offensive\_data\_extractor.cpp}}
In Cyrus2D base Version 1.0, we improved the offensive strategy, using a novel Blocking behavior, and a simplified Unmarking strategy. In this section, we will explain the improvement on the Unmarking strategy using a module called Pass Prediction.
The Pass Prediction module includes a trained DNN, that receives a state of the game, and identifies which player will be the pass receiver in that state. This module enables us to generate a tree that assigns a passer to each player in the future cycles of the game. To generate a data set for training the DNN, we employed Data Extractor module.

\subsection{Data Extractor}
As in real soccer, passing is one of the possible actions that can lead the ball to the goal. Predicting the pass target player, from the point of view of the ball possessor, has many benefits in defensive and offensive algorithms.
In this paper, the ball possessor is the player who can kick the ball in the current cycle or receive the ball in the future cycles.

To predict the behavior of (our) ball possessor, we were required to create the dataset of game states from this player point of view. For this purpose, we embedded a Data Extractor module in each one of the players and then we recorded the features of game states and their corresponding label. The label shows the uniform number of the player who is target of the best pass\cite{cyrus21,cyruschampion,cyrussamposiom}.


To generate a data set, our player (ball holder) feeds the state of the game and its selected pass receiver uniform number to the Data Extractor  when the ball is in its kickable area. After that it saves the features and the label in a CSV file. Later this dataset will be used to train the Pass Prediction model.


\subsection{Unmarking Strategy with help of Pass Prediction  Module}
\begin{figure}
    \centering
    \includegraphics[scale=0.20]{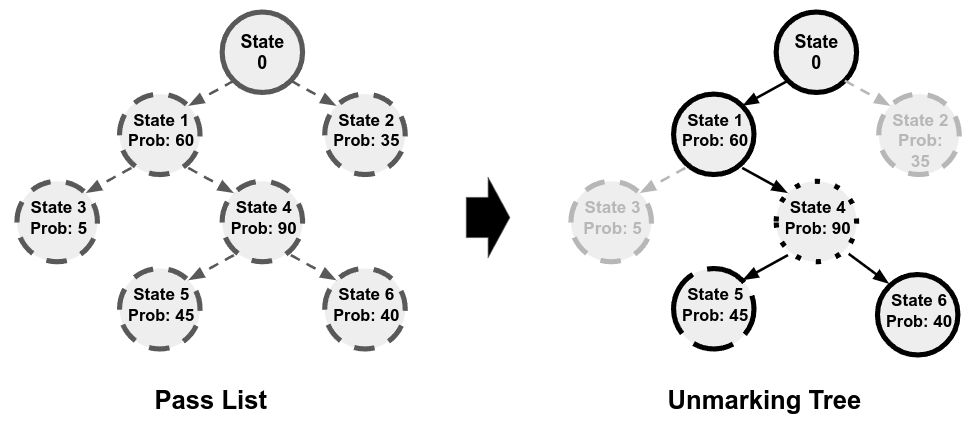} \\
    \caption{Overview of the Unmark Decisioning algorithm. The left tree shows the result states and their points. The Bold circles (full, dotted, and long dashed) in the right tree present the selected best nodes from the Candidate list. The dashed circle one indicates the node that the unmarker is the ball owner in its state, and its parent node is the dotted circle.}
    \label{fig:unmarking}
\end{figure}

To improve the Unmarking Strategy and sketching the flow of the game, each player of Cyrus2D base tries to simulate a tree that includes the probable passes and their outcoming states. Each node of the tree contains a state of the game where one of our players is the ball owner in that state. The edges from the current state shows a probable pass in the future. The root node of the tree is the first state of the game where one of our players can kick the ball. A player can not be ball possessor in more than one node of the tree. To create the tree, the unmarker feeds the state of the root node to the Pass Prediction module, and receives the probability of players for receiving a pass. This module includes a trained DNN that can receives features of the game generated by Data Extractor and gives probability of players to receive a pass in the given state. In the next step, the unmarker selects two passes with the maximum probability higher than a limit and inserts them into a list called Pass List. Then after, it pops the pass with highest probability from the Pass List. Next, it simulate the outcome that it send the outcome state to Pass Prediction module and eventually insert best passes from the outcome of Pass Prediction module. This procedure continues until the number of tree nodes is equal to ten or there is not any pass in the Pass List.
Figure \ref{fig:unmarking} shows an overview of the Unmarking Decisioning.
After termination of this procedure, the umarker agent looks for its corresponding node in the tree, then it chooses the parent player as the ”Passer” for the unmarking procedure in order to receive the pass from that player in the future.



\section{Results}
\subsection{Training DNN for Cyrus2D v1.1 \protect\footnote{All scripts for training are available in scripts/training\_unmark}}
For generating a data-set for training the "Pass Prediction DNN", we ran 500 games against Helios Base v3.11 and newest version, Gliders2D base v1.6 and v2.6, Cyurs 2021, Helios 2021, and YuShan 2021. We obtained total 1,429,032 data instances. We split them into two subsets, 85\% for training and 15\% for testing. The prediction model (DNN) has three layers of 128, 64, 32 and 11 neurons, with RELU activation function and a softmax function at the last layer. The validation accuracy of the trained neural network on the test data was 68.1\%. We used Python TensorFlow Keras library\cite{tensor} for training the model, and we implemented a library called CppDnn\cite{cppdnn} to use the trained model in C++. The CppDnn is a C++ library powered by Eigen\cite{eigen}; this library creates a deep neural network model by reading the weights of a trained DNN model.

To evaluate the impact of the implemented features and algorithms and comparing the Cyrus2D base with the HELIOS base(Ag) and Gliders2d Base(G2D), we ran X-number games between two versions of HELIOS base(3.1.1 / newest), two versions of Gliders2d Base (1.6 / 2.6) , six versions of Cyrus2D base (C2D0 = Cyrus2D zero, C2DB = Cyrus2D zero base with Blocking Strategy, C2DR = Cyrus2D zero base with ORE, C2DU = Cyrus2D zero base with UnMarking Strategy, C2DV1.0 = Cyrus2D version one with all of the previous features and C2DV1.1 = V1.0 with pass prediction) against Helios bases\cite{agent2d}, Glider2d Base(v2.6) and three of the best teams in RoboCup (Helios2021 \cite{hel21}, YuShan\cite{yushan21}, Cyrus2021 \cite{cyrus21}).

Table \ref{tab:WinRate} shows the expected winning rate of all version of Cyrus against opponent teams. The winning rate is calculated by $num\_wins / ( num\_games - num\_draws)$. Tables \ref{tab:goal_scored} presents the average number of our scored goals and conceded goals respectively.

The results demonstrate Cyrus2D base v1.1 prevalence over other released bases. For instance the Cyrus2D base wins Helios and Gliders2D bases in more than 99\% and 84\% of games respectively. The average win-rate of Cyrus2D against best three RoboCup teams is 3.76 (0.2\% to 3.8\%) percent higher than the winning rate of Helios base against those teams, and 2.86 (2.9\% to 3.8\%) percent higher than Gliders2D base.

\begin{table}[t]
    \centering
    \caption{Win Rate}
    \label{tab:WinRate}
    \begin{tabular}{l|l|l|l|l|l|l|l}
    Team   & 
    \rotatebox{60}{H2D 3.11} &
    \rotatebox{60}{H2D new} &
    \rotatebox{60}{G2D 2.6} &
    \rotatebox{60}{Cyrus21} &
    \rotatebox{60}{Helios21} &
    \rotatebox{60}{YuShan21} &
    Average \\ \hline
H2D 3.11	& --	& 26.2	& 3.5	& 0.0	& 0.0	& 0.0	& 4.9 \\
H2D new	& 73.8	& --	& 9.8	& 0.2	& 0.0	& 0.0	& 14.0 \\
G2D 1.6	& 95.5	& 85.4	& 30.3	& 0.4	& 0.0	& 0.2	& 35.3 \\
G2D 2.6	& 96.5	& 90.2	& --	& 1.9	& 0.0	& 1.0	& 31.6 \\
C2D 0.0	& 100.0	& 98.1	& 78.5	& 7.2	& 0.0	& 4.3	& 48.0 \\
C2D B	& 99.6	& 97.2	& 77.0	& 6.4	& 0.0	& 3.6	& 47.3 \\
C2D R	& 99.4	& 97.4	& 81.0	& 7.9	& 0.2	& 6.3	& 48.7 \\
C2D U	& 100.0	& 99.0	& 80.8	& 5.6	& 0.2	& 5.8	& 48.6 \\
C2D 1.0	& 99.3	& 98.6	& 79.9	& 8.6	& 0.3	& 4.6	& 48.6 \\
C2D 1.1	& 99.8	& 99.6	& 84.1	& 5.8	& 0.8	& 4.9	& 49.1 \\
    \end{tabular}
\end{table}

\begin{table}[t]
    \centering
    \caption{Goals Scored (Goals Conceded)}
    \label{tab:goal_scored}
    \begin{tabular}{l|llllll|l}
        Team   & 
    \rotatebox{60}{H2D 3.11} &
    \rotatebox{60}{H2D new} &
    \rotatebox{60}{G2D 2.6} &
    \rotatebox{60}{Cyrus21} &
    \rotatebox{60}{Helios21} &
    \rotatebox{60}{YuShan21} & Average \\ \hline
H2D 3.11	& ---	& 1.6(2.7)	& 0.5(3.0)	& 0.2(6.2)	& 0.1(13.0)	& 0.2(7.5)	& 0.4(5.4) \\
H2D new	    & 2.7(1.6)	& ---	& 0.7(2.3)	& 0.3(5.9)	& 0.1(11.1)	& 0.3(6.4)	& 0.7(4.5) \\
G2D 1.6	    & 3.5(0.7)	& 2.6(1.0)	& 0.8(1.4)	& 0.5(5.3)	& 0.1(6.5)	& 0.2(4.8)	& 1.3(3.3) \\
G2D 2.6	    & 3.0(0.5)	& 2.3(0.7)	& ---	& 0.5(3.5)	& 0.1(5.5)	& 0.2(3.8)	& 1.0(2.3) \\
C2D 0.0	    & 4.3(0.2)	& 2.8(0.3)	& 1.1(0.4)	& 0.6(2.8)	& 0.2(3.6)	& 0.3(1.9)	& 1.6(1.5) \\
C2D B	    & 4.2(0.2)	& 2.9(0.2)	& 1.1(0.4)	& 0.6(2.7)	& 0.2(3.8)	& 0.3(2.4)	& 1.6(1.6) \\
C2D R	    & 4.1(0.2)	& 2.9(0.3)	& 1.1(0.4)	& 0.6(2.6)	& 0.2(3.8)	& 0.3(1.7)	& 1.5(1.5) \\
C2D U	    & 4.4(0.2)	& 3.2(0.3)	& 1.3(0.5)	& 0.6(2.9)	& 0.2(4.0)	& 0.4(2.1)	& 1.7(1.7) \\
C2D 1.0	    & 4.8(0.2)	& 3.6(0.2)	& 1.3(0.4)	& 0.7(2.6)	& 0.2(3.9)	& 0.4(2.5)	& 1.8(1.6) \\
C2D 1.1	    & 4.4(0.2)	& 3.2(0.2)	& 1.2(0.4)	& 0.6(2.8)	& 0.2(3.8)	& 0.3(2.3)	& 1.7(1.6) \\
    \end{tabular}
\end{table}

\section{Conclusion}
In this paper, we aimed to introduce three versions of Cyrus2D base code and their particular features. The first version of Cyrus2D base was created by combining the latest release of Helios Agent2D and Gliders2D bases. For this version, we removed some of the fine tuned parameters. In the next version, Cyrus2D v1.0, we have upgraded the Blocking, and offensive strategy by using the Offensive Risk Evaluation and unmarking behavior. In the Cyrus2D v1.1 we improved the unmarking behavior using the Pass Prediction. To evaluate the performance of Cyrus2D, we ran 500 games against Gliders2D, Helios base, and best three teams in RoboCup 2021. The obtained results shows significant improvement on win-rate, scored goals and conceded goals. For our future work, we are planning to enhance the Cyrus2D base in terms of chain action movement prediction, and marking by using multi-agent decision-making.

\subsubsection*{Acknowledgements}
We acknowledge the support of the Natural Sciences and Engineering Research Council of Canada (NSERC). We thank the HELIOS and Gliders teams for their code bases and extraordinary contributions to the SS2D league.

%
%
%

\end{document}